\documentclass[runningheads]{llncs}

\usepackage{graphicx}
\usepackage{tikz}
\usepackage{comment}
\usepackage{amsthm}
\usepackage{amsmath,amssymb} % define this before the line numbering.
\usepackage{color}
\usepackage{algorithm}
\usepackage{algorithmicx}
\usepackage{algpseudocode}
\usepackage{subcaption}
\usepackage{multirow}
\usepackage{diagbox}
\usepackage{anyfontsize}

\renewcommand{\mathbf}{\boldsymbol}

\def\x{\mathbf{x}}

\def\Y{\mathbf{Y}}
\def\y{\mathbf{y}}
\def\S{\mathbf{S}}

\def\A{\mathbf{A}}
\def\a{\mathbf{a}}
\def\F{\mathbf{F}}
\def\f{\mathbf{f}}
\def\P{\mathbf{P}}
\def\p{\mathbf{p}}
\def\D{\mathbf{D}}
\def\dd{\mathbf{d}}
\def\I{\mathbf{I}}

\DeclareMathOperator*{\argmax}{arg\,max}

\usepackage{dsfont}
\newcommand{\mathbbm}[1]{{\mathds{#1}}}
\newtheorem{prop}{Proposition}

\begin{document}
% \linenumbers
\pagestyle{headings}
\mainmatter
\def\ECCVSubNumber{2763}  % Insert your submission number here

\title{Label Propagation with Augmented Anchors: \\ A Simple Semi-Supervised Learning baseline for Unsupervised Domain Adaptation} % Replace with your title

%\begin{comment}
%\titlerunning{ECCV-20 submission ID \ECCVSubNumber} 
%\authorrunning{ECCV-20 submission ID \ECCVSubNumber} 
%\author{Anonymous ECCV submission}
%\institute{Paper ID \ECCVSubNumber}
%\end{comment}
%******************

%\begin{comment}
\titlerunning{Label Propagation with Augmented Anchors}
\author{Yabin Zhang\inst{123}\and
Bin Deng\inst{12} \and
Kui Jia\inst{12} \and
Lei Zhang\inst{34}}
\authorrunning{Y. Zhang et al.}
\institute{South China University of Technology, Guangzhou, China \and
	Pazhou Lab, Guangzhou, China  \and
	DAMO Academy, Alibaba Group \and 
	Dept. of Computing, The Hong Kong Polytechnic University \\
	\email{zhang.yabin@mail.scut.edu.cn, bindeng.scut@gmail.com, kuijia@scut.edu.cn,  cslzhang@comp.polyu.edu.hk}\\
}

%\institute{South China University of Technology \and
%DAMO Academy, Alibaba Group
%\end{comment}
%******************
\maketitle

\begin{abstract}
Motivated by the problem relatedness between unsupervised domain adaptation (UDA) and semi-supervised learning (SSL), many state-of-the-art UDA methods adopt SSL principles (e.g., the cluster assumption) as their learning ingredients. However, they tend to overlook the very domain-shift nature of UDA. In this work, we take a step further to study the proper extensions of SSL techniques for UDA. Taking the algorithm of label propagation (LP) as an example, we analyze the challenges of adopting LP to UDA and theoretically analyze the conditions of affinity graph/matrix construction in order to achieve better propagation of true labels to unlabeled instances. Our analysis suggests a new algorithm of Label Propagation with Augmented Anchors (A$^2$LP), which could potentially improve LP via generation of unlabeled virtual instances (i.e., the augmented anchors) with high-confidence label predictions. To make the proposed A$^2$LP useful for UDA, we propose empirical schemes to generate such virtual instances. The proposed schemes also tackle the domain-shift challenge of UDA by alternating between pseudo labeling via A$^2$LP and domain-invariant feature learning. Experiments show that such a simple SSL extension improves over representative UDA methods of domain-invariant feature learning, and could empower two state-of-the-art methods on benchmark UDA datasets. Our results show the value of further investigation on SSL techniques for UDA problems.
\end{abstract}

\section{Introduction}
\label{SecIntro}

As a specific setting of transfer learning \cite{transfer_survey}, unsupervised domain adaptation (UDA) is to predict labels of given instances on a target domain, by learning classification models assisted with labeled data on a source domain that has a different distribution from the target one. Impressive results have been achieved by learning domain-invariant features \cite{ddc,dan,wmmd}, especially the recent ones based on adversarial training of deep networks \cite{dann,domain_confusion,mcd,symnets}. These methods are primarily motivated by the classical UDA theories \cite{ben2007analysis,ben2010theory,mansour2009,multiclass_da} that specify the success conditions of domain adaptation, where domain divergences induced by hypothesis space of classifiers are typically involved.

While a main focus of these methods is on designing algorithms to learn domain-invariant features, they largely overlook a UDA nature that shares the same property with the related problem of semi-supervised learning (SSL) --- both UDA and SSL argue for a principle that the (unlabeled) instances of interest satisfy basic assumptions (e.g., the cluster assumption \cite{chapelle2005semi}), although in SSL, the unlabeled instances follow the same distribution as that of the labeled ones. Given the advantages of SSL methods over models trained with labeled data only \cite{chapelle2009semi}, it is natural to apply the SSL techniques to domain-invariant features learned by seminal UDA methods \cite{dan,dann} so as to boost the performance further. We note that ideal domain alignment can hardly be achieved in practice. Although state-of-the-art results have already been achieved by the combination of vanilla SSL techniques and domain-invariant feature learning \cite{hou2016unsupervised,ding2018graph,li2019locality,rtn,symnets,srdc}, they typically neglect the issue that SSL methods are designed for data of the same domain, and their direct use for data with shifted distributions (e.g., in UDA tasks) could result in deteriorated performance.

To this end, we investigate how to extend SSL techniques for UDA problems. Take the SSL method of label propagation (LP) \cite{zhou2004learning} as an example. When there exists such a shift of distributions, edges of an LP graph constructed by affinity relations of data instances could be of low reliability, thus preventing its direct use in UDA problems.  To tackle the issue, we analyze in this paper the conditions of the affinity graph (and the corresponding affinity matrix) for better propagation of true labels to unlabeled instances. Our analysis suggests a new algorithm of Label Propagation with Augmented Anchors (A$^2$LP), which could potentially improve LP via generation of unlabeled virtual instances (i.e., the augmented anchors) with high-confidence label predictions. To make the proposed A$^2$LP particularly useful for UDA, we generate such virtual instances via a weighted combination of unlabeled target instances, using weights computed by the entropy of their propagated soft cluster assignments, considering that instances of low entropy are more confident in terms of their predicted labels. We iteratively do the steps of (1) using A$^2$LP to get pseudo labels of target instances, and (2) learning domain-invariant features with the obtained pseudo-labeled target instances and labeled source ones, where the second step, in turn, improves the quality of pseudo labels of target instances. Experiments on benchmark UDA datasets show that our proposed A$^2$LP significantly improves over the LP algorithm, and alternating steps of A$^2$LP and domain-invariant feature learning give state-of-the-art results. We finally summarize our main contributions as follows.

\begin{itemize}
	\item Motivated by the relatedness between SSL and UDA, we study in this paper the technical challenge that prevents the direct use of graph-based SSL methods in UDA problems. We analyze the conditions of the affinity graph/matrix construction for better propagation of true labels to unlabeled instances, which suggests a new algorithm of A$^2$LP. A$^2$LP could potentially improve LP via generation of unlabeled virtual instances (i.e., the augmented anchors) with high-confidence label predictions.
	
	\item To make the proposed A$^2$LP useful for UDA, we generate virtual instances as augmented anchors via a weighted combination of unlabeled target instances, where weights are computed based on the entropy of propagated soft cluster assignments of target instances. Our A$^2$LP based UDA method alternates in obtaining pseudo labels of target instances via A$^2$LP, and using the obtained pseudo-labeled target instances, together with the labeled source ones, to learn domain-invariant features. The second step is expected to enhance the quality of pseudo labels of target instances.
	
	\item We conduct careful ablation studies to investigate the influence of graph structure on the results of A$^2$LP. Empirical evidences on benchmark UDA datasets show that our proposed A$^2$LP significantly improves over the original LP, and the alternating steps of A$^2$LP and domain-invariant feature learning give state-of-the-art results, confirming the value of further investigating the SSL techniques for UDA problems. The codes are available at \url{https://github.com/YBZh/Label-Propagation-with-Augmented-Anchors} .
\end{itemize}

\section{Related works}
\label{SecLiterature}

In this section, we briefly review the UDA methods, especially these \cite{asy_tri,symnets,rtn,dirt,kumar2018co,french2018selfensembling,hou2016unsupervised,ding2018graph,li2019locality} involving SSL principles as their learning ingredients, and the recent works \cite{zhu2002learning,szummer2002partially,zhou2004learning,belkin2004regularization,delalleau2005efficient,joachims2003transductive} on the LP technique.

\paragraph{\textbf{Unsupervised domain adaptation}}  Motivated by the theoretical bound proposed in \cite{ben2007analysis,ben2010theory,multiclass_da}, the dominant UDA methods target at minimizing the discrepancy between the two domains, which is measured by various statistic distances, such as Maximum Mean Discrepancy (MMD) \cite{dan}, Jensen-Shannon divergence \cite{dann} and Wasserstein distance \cite{wasserstein_aaai}. They assume that once the domain discrepancy is minimized, the classifier trained on source data only can also perform well on the target ones. Given the advantages of SSL methods over models trained with labeled data only \cite{chapelle2009semi}, it is natural to apply SSL techniques on domain-invariant features to boost the results further. Recently, state-of-the-art results are achieved by involving the SSL principles in UDA, although they may not have emphasized this point explicitly. Based on the cluster assumption, entropy regularization \cite{SSLByEM} is adopted in UDA methods \cite{dirt,rtn,symnets,kumar2018co} to encourage low density separation of category decision boundaries, which is typically used in conjunction with the virtual adversarial training \cite{vat} to incorporate the locally-Lipschitz constraint. The vanilla LP method \cite{zhu2003semi} is adopted in \cite{hou2016unsupervised,ding2018graph,li2019locality} together with the learning of domain-invariant features. Based on the mean teacher model of \cite{tarvainen2017mean}, a self-ensembling (SE) algorithm \cite{french2018selfensembling} is proposed to penalize the prediction differences between student and teacher networks for the same input target instance. Inspired by the tri-training \cite{zhou2005tri}, three task classifiers are asymmetrically used in \cite{asy_tri}. However, taking the comparable LP-based methods \cite{hou2016unsupervised,ding2018graph,li2019locality} as an example, they adopt the vanilla LP algorithm directly with no consideration of the UDA nature of domain shift. By contrast, we analyze the challenges of adopting LP in UDA, theoretically characterize the conditions of potential improvement (cf. \textbf{Proposition} \ref{PROP-ACC}), and accordingly propose the algorithmic extension of LP for UDA. Such a simple algorithmic extension improves the results dramatically on benchmark UDA datasets.  

\paragraph{\textbf{Label propagation}}
The LP algorithm is based on a graph whose nodes are data instances (labeled and unlabeled), and edges indicate the similarities between instances. The labels of labeled data can propagate through the edges in order to label all nodes. Following the above principle, a series of LP algorithms \cite{zhu2002learning,szummer2002partially,zhou2004learning} and the graph regularization methods \cite{belkin2004regularization,delalleau2005efficient,joachims2003transductive} have been proposed for the SSL problems. Recently, Iscen \emph{et al.} \cite{iscen2019label} revisit the LP algorithm for SSL problems with the iterative strategy of pseudo labeling and network retraining. Liu \emph{et al.} \cite{liu2018learning} study the LP algorithm for few-shot learning. Unlike them, we investigate the LP algorithm for UDA problems and alleviate the performance deterioration brought by domain shift via the introduction of virtual instances as well as the domain-invariant feature learning.

\section{Semi-supervised learning and unsupervised domain adaptation}
\label{SecMethod}

Given data sets $X_L = \{\x_1,...,\x_l\}$ and $X_U = \{\x_{l+1},...,\x_{n}\}$ with each $\x_i \in \mathcal{X}$, the first $l$ instances have labels $Y_L = \{y_1,...,y_l\}$ with each $y_i \in \mathcal{Y} = \{1,...,K\}$ and the remaining $n - l$ instances are unlabeled. We also write them collectively as $X = \{X_L, X_U\}$. The goal of both SSL and UDA is to predict the labels of the unlabeled instances in $X_U$ \footnote{We formulate in this paper both the SSL and UDA under the transductive learning setting \cite{chapelle2009semi}.}. In UDA, the labeled data in $X_L$ and unlabeled data in $X_U$ are drawn from two different distributions of the source one $\mathcal{D}_s$ and the target one $\mathcal{D}_t$. Differently, in SSL, the source and target distributions are assumed to be the same, i.e., $\mathcal{D}_s = \mathcal{D}_t$.

\subsection{Semi-supervised learning preliminaries}

%First present the general/meta learning objective of SSL that is composed of a supervised training loss and a regularizer. Refer to equation (6) in \cite{SSLByEM}, equation (3) in \cite{ZhuSSLSurvey08}, and equation (4) in \cite{WestonSSEmbedding} for how to formulate this meta objective --- note that this meta objective should be able to be instantiated as the three specific objectives. Note also that for these different instantiations, some of the regularizers involve unlabeled data only (methods discussed in section 5 in the survey \cite{ZhuSSLSurvey08}), and others involve both labeled and unlabeled data (graph based methods discussed in section 6 in the survey \cite{ZhuSSLSurvey08})

We denote $\psi: \mathcal{X}\rightarrow \mathbb{R}^K$ as the mapping function parameterized by $\theta=\{\theta_e,\theta_c\}$, where $\theta_e$ indicates the parameters of a feature extractor $\phi: \mathcal{X}\rightarrow \mathbb{R}^d$ and $\theta_c$ indicates the parameters of a classifier $f: \mathbb{R}^d\rightarrow \mathbb{R}^K$. Let $\mathcal{P}$ denote the set of $n\times K$ probability matrices. A matrix $\P = [\p_1^T;...;\p_n^T] \in \mathcal{P}$ corresponds to a classification on the dataset $X = \{X_L, X_U\}$ by labeling each instance $\x_i$ as a label $\hat{y}_i =\arg\max_j\p_{ij}$. Each $\p_i \in [0,1]^K$ indicates classification probabilities of the instance $\x_i$ to $K$ classes.

The general goal of SSL can be stated as finding $\P$ by minimizing the following meta objective:
\begin{equation}\label{EqnMetaSSLObj}
\mathcal{Q}(\P) = \mathcal{L}(X_L, Y_L; \P) + \lambda \mathcal{R}(X; \P),
\end{equation}
where $\mathcal{L}$ represents the supervised loss term that applies only to the labeled data, and $\mathcal{R}$ is the regularizer with $\lambda$ as a trade-off parameter. The purpose of regularizer $\mathcal{R}$ is to make the learning decision to satisfy the underlying assumptions of SSL, including the smoothness, cluster, and manifold assumptions \cite{chapelle2009semi}.

For SSL methods based on cluster assumption (e.g., low density separation \cite{chapelle2005semi}), their regularizers are concerned with unlabeled data only. As such, they are more amenable to be used in UDA problems, since the domain shift is not an issue to be taken into account. A prominent example of SSL regularizer is the entropy minimization (EM) \cite{SSLByEM}, whose use in UDA can be instantiated as:
\begin{equation}\label{EqnEMSSLObj}
\begin{gathered}
\mathcal{Q}(\P) = \underbrace{\sum_{i=1}^{l}\ell(\p_i, y_i)}_{\mathcal{L}(X_L, Y_L; \P)} + \lambda\underbrace{\sum_{i=l+1}^{n}\sum_{j=1}^{K}-\p_{ij}\log \p_{ij}}_{\mathcal{R}(X; \P)} , \\
\end{gathered}
\end{equation}
where $\ell$ represents a typical loss function (e.g., cross-entropy loss). Objectives similar to (\ref{EqnEMSSLObj}) are widely used in UDA methods, together with other useful ingredients such as adversarial learning of aligned features \cite{rtn,symnets,he2018adaptive,dirt}.

\subsection{From graph-based semi-supervised learning to unsupervised domain adaptation}
%Instantiate the objective (\ref{EqnMetaSSLObj}) as an objective for graph based SSL methods that are based on local (and global) smoothness, and thus their regularizers are concerned with both labeled and unlabeled data. Point out that for graph based methods, their ways of affinity matrix construction are more important to achieve good results. Here may choose to instantiate this objective as the LP based SSL method ((\ref{EqnLPSSLObj})) (i.e., some objective similar to equation (4) in \cite{WestonSSEmbedding} or equation (4) in \cite{ZhouLP}),

\begin{figure}[htb]
		\begin{minipage}[t]{0.31\linewidth}
			\centering
			\includegraphics[width=0.94\linewidth] {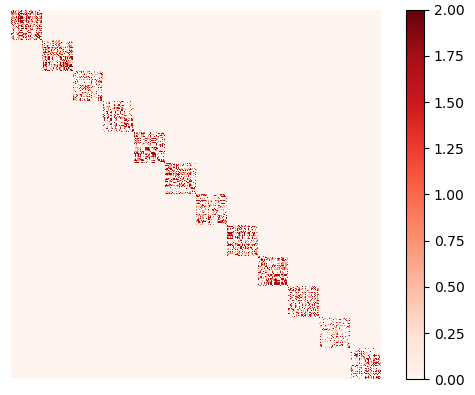}
			\subcaption{SSL (93.5\%)}
			\label{Fig:ssl}
	\end{minipage}
	\hfill
		\begin{minipage}[t]{0.31\linewidth}
			\centering
			\includegraphics[width=0.94\linewidth] {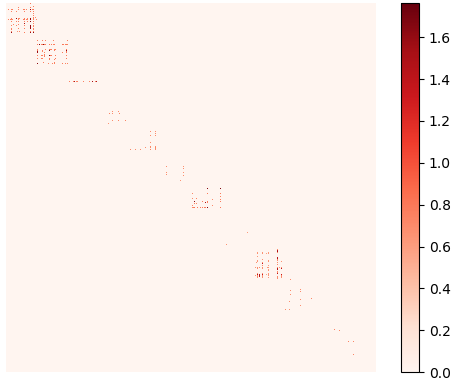}
			\subcaption{UDA (64.8\%)}
			\label{Fig:da}
	\end{minipage}
	\hfill
		\begin{minipage}[t]{0.31\linewidth}
			\centering
			\includegraphics[width=0.94\linewidth] {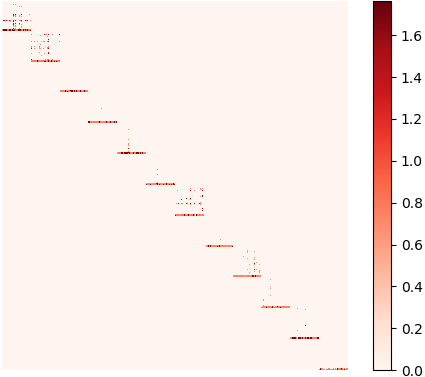}
			\subcaption{\scriptsize UDA with Anchors (79.5\%)}
			\label{Fig:daaa}
	\end{minipage}
	\caption{Visualization of sub-affinity matrices for the settings of (a) SSL, (b) UDA, and (c) UDA with augmented anchors, and their corresponding classification results via the LP. The row-wise and column-wise elements are the unlabeled and labeled instances, respectively. For illustration purposes, we keep elements connecting instances of the same class unchanged, set the others to zero, and sort all instances in the category order using the ground truth category of all data. As we can see, the augmented anchors present better connections with unlabeled target instances compared to the labeled source instances in UDA.}
	\label{Fig:feature_matrix}
\end{figure}

Different from the above EM like methods, the graph-based SSL methods that are based on local (and global) smoothness rely on the geometry of the data, and thus their regularizers are concerned with both labeled and unlabeled instances. The key of graph-based methods is to build a graph whose nodes are data instances (labeled and unlabeled), and edges represent similarities between instances. Such a graph is represented by the affinity matrix $\A \in \mathbb{R}_+^{n\times n}$, whose elements $\a_{ij}$ are non-negative pairwise similarities between instances $\x_i$ and $\x_j$. Here, we choose the LP algorithm \cite{zhou2004learning} as an instance for exploiting the advantages of graph-based SSL methods. Denote $\Y = [\y_1^T;...;\y_n^T] \in \mathcal{P}$ as the label matrix with $\y_{ij} = 1$ if $\x_i$ is labeled as $y_i = j$ and $\y_{ij} = 0$ otherwise. The goal of LP is to find a $\F = [\f_1^T;...;\f_n^T]\in\mathbb{R}_+^{n\times K}$ by minimizing
%The meta objective of (\ref{EqnMetaSSLObj}) can be instantiated to LP as
\begin{equation}\label{EqnLPSSLObj}
\mathcal{Q}(\F) = \underbrace{\sum_{i=1}^n\|\f_i - \y_i\|^2}_{\mathcal{L}(X_L, Y_L; \F)} + \lambda \underbrace{\sum_{i,j}^{n}\a_{ij}\|\frac{\f_i}{\sqrt{\dd_{ii}}} - \frac{\f_j}{\sqrt{\dd_{jj}}}\|^2}_{\mathcal{R}(X; \F)},
\end{equation}
and then the resulting probability matrix $\P$ is given by $\p_{ij} = \f_{ij}/\sum_j\f_{ij}$, where $\D\in \mathbb{R}_+^{n \times n}$ is a diagonal matrix with its $(i,i)$-element $\dd_{ii}$ equal to the sum of the $i$-th row of $\A$. From the above optimization objective, we can easily see that a good affinity matrix is the key success factor of the LP algorithm. So, the straightforward question is that what is the good affinity matrix? As an analysis, we assume the true label of each data instance $\x_i$ is $y_i$, then the regularizer of the objective (\ref{EqnLPSSLObj}) can be decomposed as:
\begin{equation}
\begin{aligned}\label{EqnLPSSLSubObjective}
\mathcal{R}(X; \F) = \sum_{y_i = y_j}\a_{ij}\|\frac{\f_i}{\sqrt{\dd_{ii}}} - \frac{\f_j}{\sqrt{\dd_{jj}}}\|^2 \\
+ \sum_{y_i \neq y_j}\a_{ij}\|\frac{\f_i}{\sqrt{\dd_{ii}}} - \frac{\f_j}{\sqrt{\dd_{jj}}}\|^2.
\end{aligned}
\end{equation}
Obviously, a good affinity matrix should make its element $\a_{ij}$ as large as possible if instances $\x_i$ and $\x_j$ are in the same class, and at the same time make those $\a_{ij}$ as small as possible otherwise. Therefore, it is rather easy to construct such a good affinity matrix in the SSL setting where the all data are drawn from the same underlying distribution. However, in UDA, due to the domain shift between labeled and unlabeled data, those values of elements $\a_{ij}$ of the same class pairs between labeled and unlabeled instances would be significantly reduced, which would prevent its use in the UDA problems as illustrated in Figure \ref{Fig:feature_matrix}.

\begin{figure}[htb]
	\centering
	\includegraphics[width=0.98\linewidth] {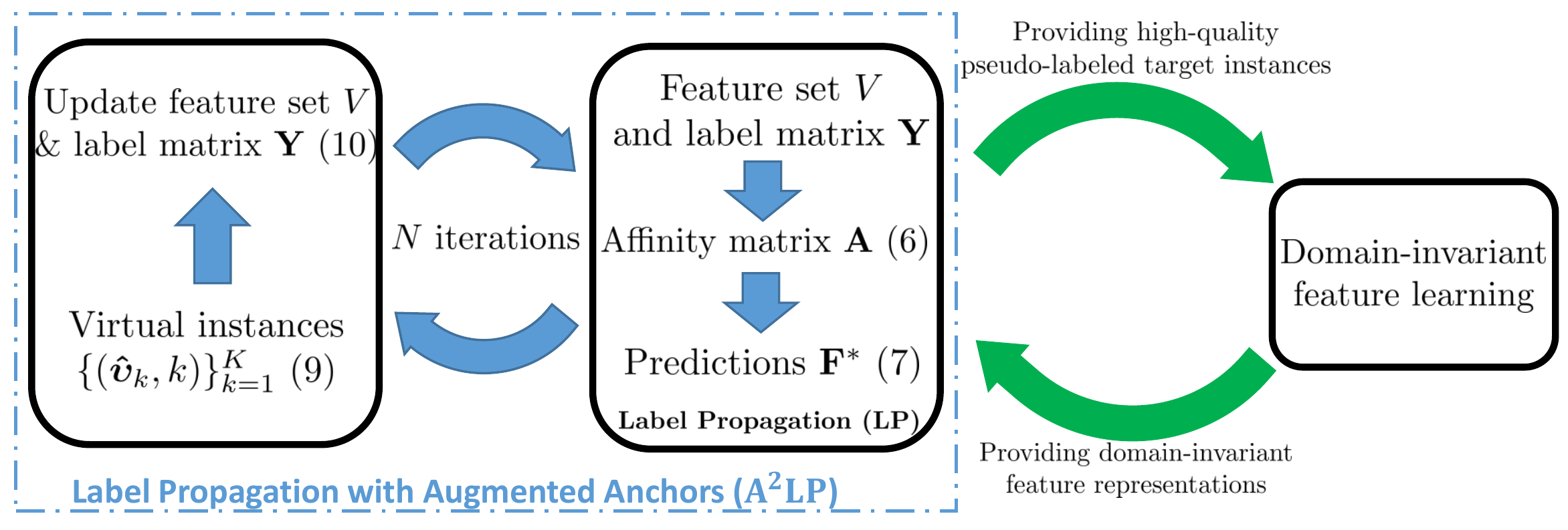}
	\caption{An illustration of the overall framework of alternating steps of pseudo labeling via A$^2$LP and domain-invariant feature learning. The dashed line rectangle illustrates the algorithm of A$^2$LP, where we iteratively do the steps of (1) augmenting the feature set $V$ and label matrix $\mathbf{Y}$ with the generated virtual instances and (2) generating virtual instances by the LP algorithm based on the updated feature set $V$ and label matrix $\mathbf{Y}$.}
	\label{Fig:framework}
\end{figure}

\section{Label propagation with augmented anchors} \label{SecA2LP}
In this section, we first analyze conditions of the corresponding affinity matrix for better propagation of true labels to unlabeled instances, which motivate our proposed A$^2$LP algorithm.
Let $Acc$ be the classification accuracy in $X_U$ by the solution of the LP (Equ. (\ref{EqnLPSSLObj})), i.e.,
\begin{equation}\label{Equ:acc}
Acc := \frac{|\{\x_i\in X_U:\hat{y}_i = y_i\}|}{| X_U |},
\end{equation}
where $\hat{y}_i = \argmax_j \f^*_{ij}$ with $\F^* = [{\f^*_1}^T;...;{\f^*_n}^T]$ the solution of Equ. (\ref{EqnLPSSLObj}).

\begin{prop} 
	\label{PROP-ACC}
	Assume the data satisfy the ideal cluster assumption, i.e., $\a_{ij} = 0$ for all $y_i\neq y_j$. Enhancing one zero-valued element $\a_{mn}$ between a data instance $\x_m$ (labeled or unlabeled) and a labeled instance $\x_n \in X_L$ to a positive number, where $y_m = y_n$, the $Acc$ (\ref{Equ:acc}) non-decreases, and increases under the condition when originally $\hat{y}_m \neq y_m$.
\end{prop}
The proof of Proposition \ref{PROP-ACC} can be found in the appendices.

\begin{remark}\label{REMARK1}
	Under the assumption of Proposition \ref{PROP-ACC},
	%the affinity matrix can be constructed by k-nearest neighbor to approach the condition: $\a_{ij} = 0$ for those $y_i\neq y_j$. Then,
	if we can augment the labeled set $X_L$ with one virtual instance with the true label, whose neighbors are exactly the instances with the same label, then based on Proposition \ref{REMARK1}, the LP algorithm can get increasing (non-decreasing) $Acc$ (Equ. (\ref{Equ:acc})) in $X_U$.
\end{remark}

\subsection{The proposed algorithms}

%\cred{The theoretical analysis presented above suggests the algorithm of Label Propagation with Augmented Anchors ($\textrm{A}^2$LP). Key to success of the proposed algorithm is how to produce virtual instances whose distribution follows the unlabeled target data and whose assigned labels are accurate with high confidence.

%Then present our proposed $\textrm{A}^2$LP algorithm, its variants, and explain why they could work by empirically satisfying the asumptions/conditions.}

Based on the above analysis, we propose the algorithm of Label Propagation with Augmented Anchors (A$^2$LP), as illustrated in Fig. \ref{Fig:framework}. We detail the A$^2$LP method as follows.

\paragraph{\textbf{Nearest neighbor graph.}} We construct the feature set $V = \{\mathbf{v}_1, \cdots, \mathbf{v}_l, \mathbf{v}_{l+1}, \cdots,\mathbf{v}_n\}$, where $\mathbf{v}_i := \phi_{\theta_e}(\mathbf{x}_i)$. The affinity matrix $\mathbf{A} \in \mathbb{R}_+^{n\times n}$ is constructed with elements:
\begin{equation} \label{Equ:graph}
\mathbf{a}_{ij}:= \left\{ \begin{array}{cl}
\varepsilon(\mathbf{v}_i, \mathbf{v}_j), &\mathrm{if} \quad i \neq j \land \mathbf{v}_i \in  \mathrm{NN}_k(\mathbf{v}_j)  \\ 0, & otherwise
\end{array}   \right.
\end{equation}
% \frac{<\mathbf{v}_i, \mathbf{v}_j>}{\|\mathbf{v}_i \| \|\mathbf{v}_j \|}  cosine similarity
where $\varepsilon(\mathbf{v}_i, \mathbf{v}_j)$ measures the non-negative similarity between $\mathbf{v}_i$ and $\mathbf{v}_j$, and $\mathrm{NN}_{k}$ denotes the set of $k$ nearest neighbors in $X$.
Then, we adopt $\mathbf{A} =\mathbf{A} + \mathbf{A}^T$ to make  $\mathbf{A}$ a symmetric non-negative adjacency matrix with zero diagonal.

\paragraph{\textbf{Label propagation.}} The closed-form solution of the objective (Equ. (\ref{EqnLPSSLObj})) of the LP algorithm is given by \cite{zhou2004learning} as
\begin{equation}\label{Eqn:closed-form-solution}
\F^* = (\I - \alpha \S)^{-1}\Y,
\end{equation}
where $\alpha = \frac{2\lambda}{2\lambda + 1}$, $\mathbf{I}$ is an identity matrix and $\S = \D^{-1/2}\A\D^{-1/2}$.

\paragraph{\textbf{LP with augmented anchors.}} Suggested by the Remark \ref{REMARK1}, we generate virtual instances via a weighted combination of unlabeled target instances, using weights computed by the entropy of their propagated soft cluster assignments, considering that instances of low entropy are more confident in terms of their predicted labels. In particular, we first obtain the pseudo labels of unlabeled target instances by solving Equ. (\ref{Eqn:closed-form-solution}), and then we assign the weight $w_i$ to each unlabeled instance $\mathbf{x}_i$ by
%we then iteratively add estimated target category centers as virtual samples to the labeled set. Although we can obtain the pseudo labels of unlabeled target samples by solving Equation (\ref{Eqn:closed-form-solution}), we do not have the same confidence for each sample. Thus, we assign weight $w_i$ to each unlabeled instance $\mathbf{x}_i$ by involving the entropy in the measurement of the confidence:
\begin{equation} \label{Equ:ins_weight}
w_i := 1 - \frac{H(\mathbf{\p}^*_i)}{\log(K)},
\end{equation}
where $H(\cdot)$ is the entropy function and $\mathbf{\p}^*_{ij} = {\mathbf{\f}}^*_{ij} / \sum_j{{\mathbf{\f}}^*_{ij}}$. We have $w_i \in [0,1]$ since $0 \leq H(\mathbf{\p}^*_i) \leq \log{K}$. The virtual instances $\{\left(\hat{\mathbf{v}}_{n+k}, k \right)\}_{k=1}^K$ can then be calculated as:
\begin{equation} \label{Equ:cate_centers}
\left(\hat{\mathbf{v}}_{n+k}, k \right) = \left(\sum_{\x_i \in X_U} \frac{\mathbbm{1}(k=\hat{y}_i) w_i \phi_{\theta_e}(\mathbf{x}_i)}{\sum_{\x_j \in X_U} \mathbbm{1}(k=\hat{y}_j) w_j}, k \right),
\end{equation}
where $\mathbbm{1}(\cdot)$ is the indicator function. The virtual instances generated by Equ. (\ref{Equ:cate_centers}) are relatively robust to the label noise and their neighbors are probably the instances of the same label due to the underlying cluster assumption.

Then, we iteratively do the steps of (1) augmenting the feature set $V$ and label matrix $\mathbf{Y}$ with the generated virtual instances and (2) generating virtual instances by the LP algorithm based on updated feature set $V$ and label matrix $\mathbf{Y}$.
The updating strategies of feature set $V$ and label matrix $\mathbf{Y}$ are as follows:
\begin{align} \label{Equ:update_strategy}
 V = V \cup \{\hat{\mathbf{v}}_{n+1}, \cdots, \hat{\mathbf{v}}_{n+K} \},  
\mathbf{Y} =  \left[ \begin{array}{c}
\mathbf{Y}  \\
\mathbf{I}  \\
\end{array}   \right] ,
n &= n + K.
\end{align}
The iterative steps empirically converge in less than $10$ iterations, as illustrated in Sec. \ref{Sec:analysis}. The implementation of our A$^2$LP is summarized in Algorithm \ref{alg:ILP} (line 2 $to$ 10).

\begin{algorithm}[t]
	\caption{Alternating steps of pseudo labeling via A$^2$LP and domain-invariant feature learning.}	
	\label{alg:ILP}
	\textbf{Input:} \\
	Labeled data: $\{X_L,Y_L\} = \{(\mathbf{x}_{1}, y_1), ..., (\mathbf{x}_{l},y_l)\}$ \\
	Unlabeled data: $X_U = \{\mathbf{x}_{l+1}, ..., \mathbf{x}_{n}\}$ \\
	Model parameters: $\theta = [\theta_e, \theta_c]$\\
	\textbf{Procedure:}
	\begin{algorithmic}[1]
		\While{Not Converge}
		\State Construct feature set $V$ and label matrix $\mathbf{Y}$;
		\For {iter $=1$ to $N$}  \Comment{Pseudo labeling via A$^2$LP}
		\State Compute affinity matrix $\mathbf{A}$ by Equ. (\ref{Equ:graph});
		\State $\mathbf{A} \leftarrow \mathbf{A} + \mathbf{A}^{T}$;
		\State $\mathbf{S} \leftarrow \mathbf{D}^{-1/2}\mathbf{A}\mathbf{D}^{-1/2}$;
		\State Get predictions $\mathbf{\F}^*$ by Equ. (\ref{Eqn:closed-form-solution});
		\State Calculate the virtual instances by Equ. (\ref{Equ:cate_centers});
		\State Update $V$ and $\mathbf{Y}$ with virtual instances by Equ. (\ref{Equ:update_strategy});
		\EndFor
		\State Remove added virtual instances, and $n = n - NK$;
		\For {iter $=1$ to $M$} \Comment{Domain-invariant feature learning}
		\State Update parameters $\theta$ by domain-invariant feature learning (e.g., \cite{semantic_align,can});
		\EndFor
		\EndWhile
	\end{algorithmic}	
\end{algorithm}

\paragraph{\textbf{Alternating steps of pseudo labeling and domain-invariant feature learning.}}

%Following the pseudo label based SSL methods \cite{}, we first introduce the alternating steps of pseudo labeling and network training with the simple cross-entropy loss as a baseline. The training loss function is as:
%\begin{align} \label{Equ:Cross_entopy}
%\nonumber	\mathcal{L}(X_L, Y_L, X_U, \hat{Y}_U; \theta) & =  \sum_{i=1}^{l} \frac{\ell(\psi_{\theta}({\mathbf{x}_i}), y_i)}{l} \\
%+ &\xi \sum_{i=l+1}^n \frac{\mathbbm{1}(w_i > \tau) w_i \ell(\psi_{\theta}(\mathbf{x}_i), \hat{y}_i)}{\sum_{j=l+1}^n \mathbbm{1}(w_j > \tau) w_j},
%\end{align}
%where $\ell$ is the cross-entropy loss function, $\tau$ is the score threshold to filter samples of low prediction confidence, $\xi$ controls the weight of loss of target data classification, and $\hat{Y}_U = \{ \hat{y}_{l+1}, \cdots, \hat{y}_{n}\}$ is the collection of pseudo labels of unlabeled set $X_U$.

Although our proposed A$^2$LP can largely alleviate the performance degradation of applying LP to UDA tasks via the introduction of virtual instances, learning domain-invariant features across labeled source data and unlabeled target data is fundamentally important, especially when the domain shift is unexpectedly large. To illustrate the advantage of our proposed A$^2$LP on generating high-quality pseudo labels of unlabeled data, and to justify the efficacy of the alternating steps of pseudo labeling via SSL methods and domain-invariant feature learning, we empower state-of-the-art UDA methods \cite{semantic_align,can} by replacing their pseudo label generators with our A$^2$LP, and keep other settings unchanged. Empirical results in Sec. \ref{Sec:results} testify the efficacy of our A$^2$LP.

\paragraph{\textbf{Time Complexity of A$^2$LP.}}
Computation of our proposed algorithm is dominated by constructing the affinity matrix (\ref{Equ:graph}) via $k$-nearest neighbor graph and solving the closed-form solution (\ref{Eqn:closed-form-solution}). Brute-force implementations of them are computationally expensive for datasets with large numbers of instances. Fortunately, the $O(n^2)$ complexity of full affinity matrix construction of the $k$-nearest neighbor graph can be largely improved via NN-Descent \cite{dong2011efficient}, giving rise to an almost linear empirical complexity of $O(n^{1.1})$. Given that the matrix $(\I - \alpha \S)$ is positive-definite, the label predictions $\F^*$ (\ref{Eqn:closed-form-solution}) can be achieved by solving the following linear system with the conjugate gradient (CG) \cite{hestenes1952methods,zhu2005semi}:
\begin{equation}\label{Eqn:closed-form-solution-cg}
(\I - \alpha \S) \F^* = \Y,
\end{equation}
which is known to be faster than the closed-form solution (\ref{Eqn:closed-form-solution}).
Empirical results in the appendices show that such accelerating strategies significantly reduce the time consumption and hardly suffer performance penalties.

%As the A$^2$LP can introduce (possibly) better pseudo labels of target instances, we train the deep network $\psi$ for discriminative feature representations with the pseudo-labeled target instances and source labeled ones. The training loss function is as:
%\begin{align} \label{Equ:Cross_entopy}
%\nonumber	\mathcal{L}(X_L, Y_L, X_U, \hat{Y}_U; \theta) & =  \sum_{i=1}^{l} \frac{\ell(\psi_{\theta}({\mathbf{x}_i}), y_i)}{l} \\
%+ &\xi \sum_{i=l+1}^n \frac{\mathbbm{1}(w_i > \tau) w_i \ell(\psi_{\theta}(\mathbf{x}_i), \hat{y}_i)}{\sum_{j=l+1}^n \mathbbm{1}(w_j > \tau) w_j},
%\end{align}
%where $\ell$ is the cross-entropy loss function, $\tau$ is the score threshold to filter samples of low prediction confidence, $\xi$ controls the weight of loss of target data classification, and $\hat{Y}_U = \{ \hat{y}_{l+1}, \cdots, \hat{y}_{n}\}$ is the collection of pseudo labels of unlabeled set $X_U$.
%Since the discriminative feature leaning of deep network would in turn benefit the construction of affinity matrix, we therefore alternate in obtaining (possibly) better pseudo labels of target instances via A$^2$LP, and using the obtained pseudo-labeled target instances (together with the labeled source ones) to train the UDA network with loss (\ref{Equ:Cross_entopy}), which is illustrated in Algorithm \ref{alg:ILP}.
%Other alternatives of the loss $\mathcal{L}$ could be that proposed in existing methods \cite{semantic_align,pan2019transferrable,can}.
%

\section{Experiments}
%
%In this section, we evaluate our A$^2$LP on three UDA benchmark datasets.
%
%\subsection{Experiments Setup} \label{Sec:dataset_imple}
\noindent \textbf{Office-31} \cite{saenko2010adapting} is a standard UDA dataset including three diverse domains: Amazon (\textbf{A}) from Amazon website, Webcam (\textbf{W}) by web camera, and DSLR (\textbf{D}) by digital SLR camera. There are $4,110$ images of $31$ categories shared across three domains. %, with $2,817$ images in \textbf{A} domain, $795$ images in \textbf{W} domain, and $498$ images in \textbf{D} domain.
\noindent \textbf{ImageCLEF-DA} \cite{ImageCLEFDA} is a balanced dataset containing three domains: Caltech-256 (\textbf{C}), ImageNet ILSVRC 2012 (\textbf{I}), and Pascal VOC 2012 (\textbf{P}). There are $12$ categories and $600$ images in each domain.
\noindent \textbf{VisDA-2017} \cite{peng2017visda} is a dataset with large domain shift from the synthetic data (\textbf{Syn.}) to real images (\textbf{Real}). There are about $280$K images across 12 categories.

We implement our A$^2$LP based on PyTorch. We adopt the ResNet \cite{resnet} pre-trained on the ImageNet dataset \cite{deng2009imagenet} excluding the last fully connected (FC) layer as the feature extractor $\phi$. In the alternating training step, we fine-tune the feature extractor $\phi$ and train a classifier $f$ of one FC layer from scratch. We update all parameters by stochastic gradient descent with momentum of 0.9, and the learning rate of the classifier $f$ is $10$ times that of the feature extractor $\phi$.
We employ the annealing strategy of learning rate \cite{dann} following $\eta_p = \frac{\eta_0}{(1+\mu p)^{\beta}}$, where $p$ is the process of training iterations linearly changing from $0$ to $1$, $\eta_0=0.01$ and $\mu=10$. Following \cite{can}, we set $\beta=0.75$ for datasets of Office-31 \cite{saenko2010adapting} and ImageCLEF-DA \cite{ImageCLEFDA}, while for VisDA-2017 dataset, $\beta=2.25$.
We adopt the cosine similarity, i.e., $\varepsilon(\mathbf{v}_i, \mathbf{v}_j) = \frac{<\mathbf{v}_i, \mathbf{v}_j>}{\|\mathbf{v}_i \| \|\mathbf{v}_j \|}$, to construct the affinity matrix (\ref{Equ:graph}) and compare it with other two alternatives in Sec. \ref{Sec:analysis}. We empirically set $\alpha$ as $0.75$ and $0.5$ for the VisDA-2017 dataset and datasets of Office-31 and ImageCLEF-DA, respectively. We use all labeled source data and all unlabeled target data in the training process, following the standard protocols for UDA \cite{dann,dan}. For each adaptation task, we report the average classification accuracy and the standard error on three random experiments.

%\subsection{Ablation Studies} \label{Sec:ablation_exp}
\subsection{Analysis} \label{Sec:analysis}
\begin{figure}[htbp]
	\centering
	\begin{minipage}{0.48\textwidth}
		\includegraphics[width=0.98\linewidth] {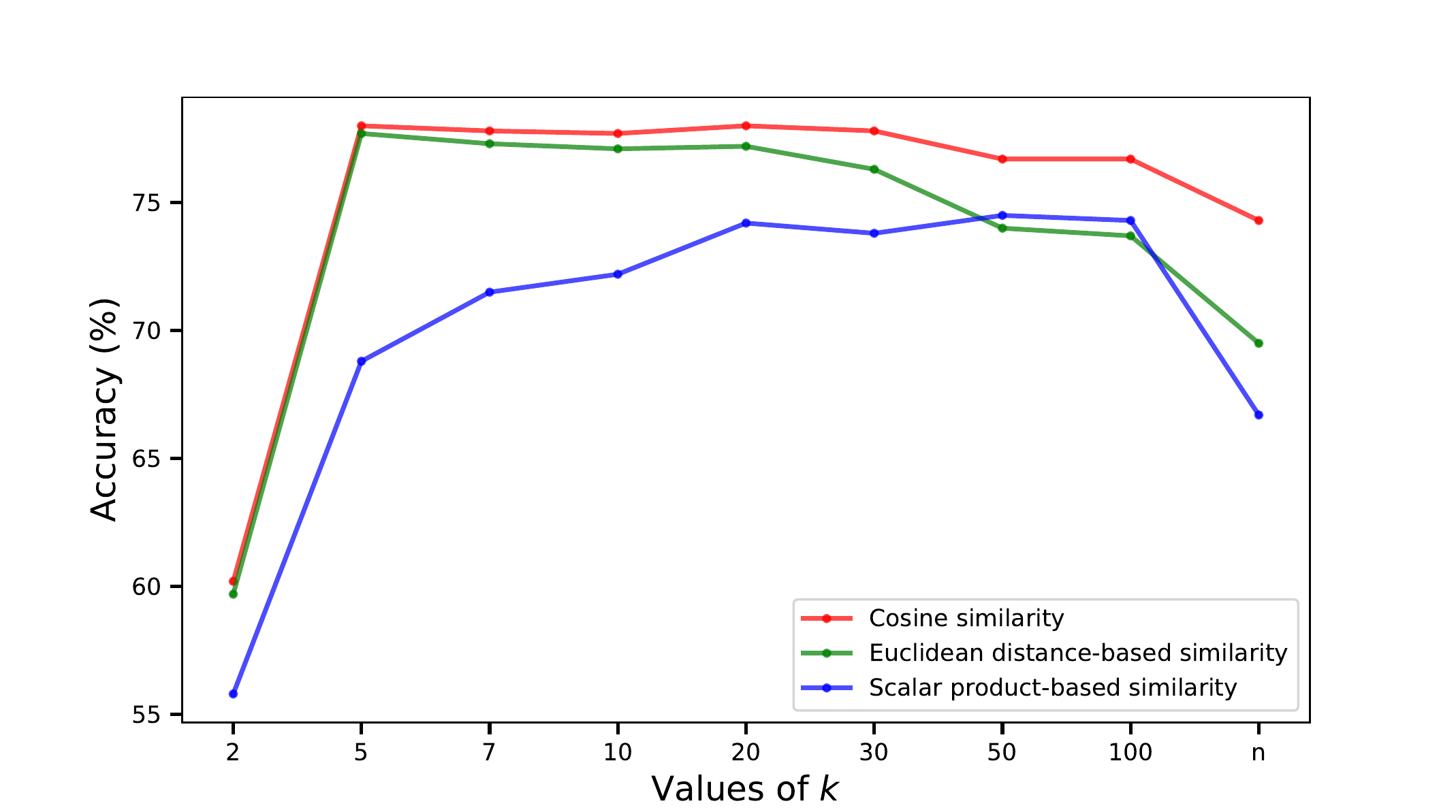}
		\caption{An illustration of the accuracy (\%) of A$^2$LP with affinity matrix constructed with different similarity measurements and different values of $k$. Results are reported on the \textbf{C} $\to$ \textbf{P} task of the ImageCLEF-DA dataset based on a 50-layer ResNet. When $k=n$, we construct the full affinity matrix as in \cite{ZhouLP}. Please refer to Sec. \ref{Sec:analysis} for the definitions of similarities.}
		\label{Fig:similarity_abla}
	\end{minipage}
	\hspace{0.2cm}
	\begin{minipage}{0.48\textwidth}
		\centering
		\includegraphics[width=0.98\linewidth] {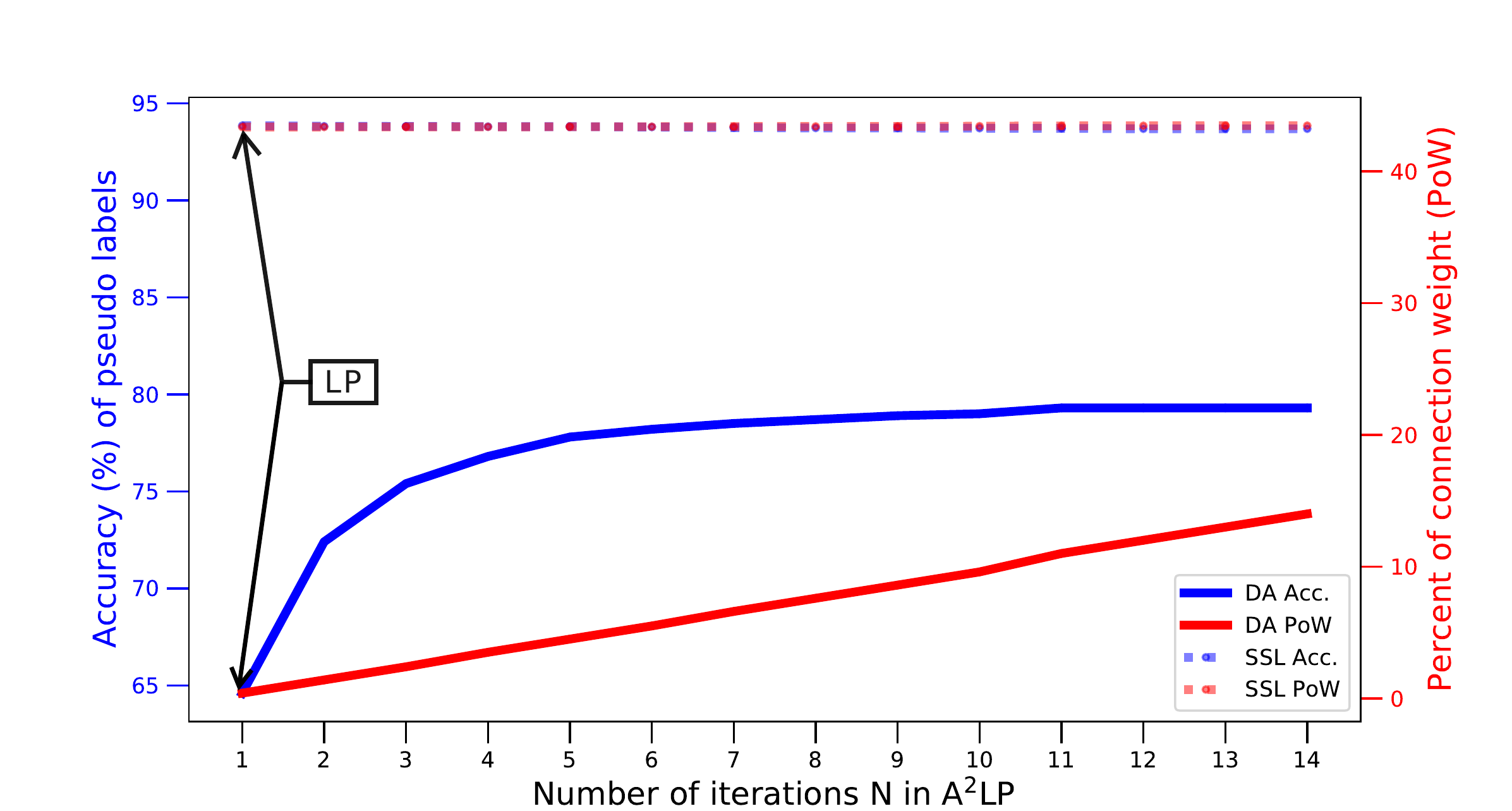}
		\caption{An illustration of the \textcolor{blue} {accuracy (\%) of pseudo labels of unlabeled instances (left y-axis)} and the \textcolor{red} {percent (\%) of connection weight (PoW) of the same class pairs between labeled and unlabeled data (right y-axis)} of our proposed A$^2$LP on the tasks of SSL and UDA. The A$^2$LP degenerates to LP \cite{zhou2004learning} when the number of iteration $N$ is set to 1. Please refer to Section \ref{Sec:analysis} for the detailed settings.}
		\label{Fig:converge_A$^2$LP}
	\end{minipage}
\end{figure}

\paragraph{\textbf{Various Similarity Metrics}}
In this section, we conduct ablative experiments on the \textbf{C} $\to$ \textbf{P} task of the ImageCLEF-DA dataset to analyze the influences of graph structures to results of A$^2$LP. To study the impact of similarity measurements, we construct the affinity matrix with other two alternative similarity measurements, namely the Euclidean distance-based similarity $\varepsilon(\mathbf{v}_i, \mathbf{v}_j) = \exp(-\| \mathbf{v}_i - \mathbf{v}_j \|^2/2)$ introduced in \cite{ZhouLP} and the scalar product-based similarity $\varepsilon(\mathbf{v}_i, \mathbf{v}_j) = \max(\mathbf{v}_i^T \mathbf{v}_j, 0)^3$ adopted in \cite{iscen2017efficient}. We also set $k$ to different values to investigate its influence.
Results are illustrated in Figure \ref{Fig:similarity_abla}. We empirically observe that results with cosine similarity are consistently better than those with the other two alternatives. We attribute the advantages of cosine similarity to the adopted FC layer-based classifier, where the cosine similarities between features and weights of the classifier dominate category predictions. The results of A$^2$LP with affinity matrix constructed by the cosine similarity are stable under a wide range of $k$ (i.e., $5\sim 30$). Results with the full affinity matrix (i.e., $k=n$) are generally lower than that with the nearest neighbor graph. We empirically set $k=20$ in the experiments for the Office-31 and ImageCLEF-DA datasets, and set $k=100$ for the VisDA-2017 dataset, where the number of instances is considerably large.

%\begin{figure}
%	\centering
%	\includegraphics[width=0.5\linewidth] {figure/ssda_convergence.pdf}
%	\caption{An illustration of the \textcolor{blue} {accuracy (\%) of pseudo labels of unlabeled instances (left y-axis)} and the \textcolor{red} {percent (\%) of connection weight (PoW) of the same class pairs between labeled and unlabeled data (right y-axis)} of our proposed A$^2$LP on the tasks of SSL and UDA. The A$^2$LP degenerates to LP \cite{zhou2004learning} when the number of iteration $N$ is set to 1. Please refer to Section \ref{Sec:analysis} for the detailed settings.}
%	\label{Fig:converge_A$^2$LP}
%\end{figure}

\paragraph{\textbf{A$^2$LP on UDA and SSL}}
In this section, we observe the behaviors of the A$^2$LP on UDA and SSL tasks. The goal of the experiment is to observe the results with augmented virtual instances in LP.  For the labeled data, we randomly sample $1000$ instances per class in the synthetic image set of the VisDA-2017 dataset. For the SSL task, we randomly sample another $1000$ instances per class in the synthetic image set of the VisDA-2017 dataset as the unlabeled data, whereas $1000$ instances are sampled randomly in each class of the real image set to construct unlabeled data in the UDA task. We denote the constructed UDA task as VisDA-2017-Small for ease of use. The mean prediction accuracy of all unlabeled instances is reported. To give insights of the different results, we illustrate the percent of connection weight (PoW) of the same class pairs between labeled and unlabeled data in the constructed k-nearest neighbor graph using ground-truth labels of unlabeled data, which is calculated as: $PoW = \frac{W_{lu}}{W_{all}}$, where $W_{lu}$ is the similarities sum of connections of the same class pairs between labeled and unlabeled data in the affinity matrix $\textbf{A}$ and $W_{all} = \sum_{i,j}\a_{ij}$.

The results are illustrated in Fig. \ref{Fig:converge_A$^2$LP}. In the UDA task, the initial PoW (i.e., N=1) is too low to enable the labels of labeled data propagate to all the unlabeled target data. As the A$^2$LP proceeds, the labeled data are augmented with virtual instances with true labels, whose neighbors involve unlabeled instances sharing the same label. Thus the PoW increases, leading to more accurate predictions of unlabeled data. In the SSL task, cluster centers of labeled and unlabeled data are positioned to be close and (statistically) in relatively dense population areas, since all data follow the same distribution. Instances close to cluster centers, including the $k$ nearest neighbors of virtual instances, are expected to be classified correctly by the LP algorithm, leading to unchanged results as the A$^2$LP proceeds. 
These observations corroborate the \textbf{Proposition} \ref{PROP-ACC} and verify the efficacy of our proposed virtual instances generation strategy for UDA.

\paragraph{\textbf{Robustness To Noise}}
We investigate the influence of the noise level of label predictions on the results of A$^2$LP. As illustrated in Table \ref{Tab:noise}, the A$^2$LP is robust to the label noise. Specifically,  as the noise level increases,  results of A2LP degrade and are worse than that of the vanilla LP when the noise level is larger than $\sim$60\%. 

\begin{table}[htb]
	\centering
	\begin{tabular}{l|cccccccc|| c}
		\hline
		Noise level (\%)        & 0     & 10    & 30    & 50 & 60 & 70 & 80  & 100 & Vanilla LP \\
		\hline
		Acc. (\%) of A$^2$LP    & 92.8  & 92.8 &92.3     & 91.8   & 91.0 & 90.7 & 90.3 & 90.0 & 91.2\\
		\hline
	\end{tabular}
	\caption{Results of A$^2$LP with different noise levels of initial label predictions on the \textbf{P} $\to$ \textbf{C} task of the ImageCLEF-DA dataset. We replace the initial label predictions from the LP (i.e., the Line 7 of the Algorithm \ref{alg:ILP}) with a manually defined setting, where the noise level of L\% indicates that the virtual instances (i.e., Equ. (9)) are calculated with unlabeled target data, L\% of which are assigned with random and wrong pseudo labels. Note that we set N=2 (cf. Line 3 of Algorithm \ref{alg:ILP}) here.}
	\label{Tab:noise}
\end{table}
\paragraph{\textbf{A$^2$LP variant}}
We propose a degenerated variant of A$^2$LP by representing the entire labeled source data with several representative surrogate instances in the A$^2$LP process, which can largely alleviate the computation cost of the LP algorithm. More specifically, we replace the features of source data $\{\mathbf{v}_{1}, \cdots, \mathbf{v}_{l} \}$ with $K$ source category centers $\{\mathbf{v}_{k} = \sum_{i=1}^l \frac{\mathbbm{1}(k=y_i) \mathbf{v}_i}{\sum_{j=1}^{l} \mathbbm{1}(k=y_j)}\}_{k=1}^K$ with category labels $\{1, \cdots, K\}$ (only the Line 2 of the Algorithm \ref{alg:ILP} is updated accordingly). As illustrated in Table \ref{Tab:variant}, the result of A$^2$LP variant is slightly lower than that of the A$^2$LP on the VisDA-2017-Small task.
Note that we only adopt the A$^2$LP variant in tasks involving the entire VisDA-2017 dataset unless otherwise specified.

\begin{table}
	\begin{minipage}{0.48\linewidth}
		\centering
		\caption{Comparison between the A$^2$LP and its degenerated variant on the VisDA-2017-Small task based on a 50-layer ResNet.}
		\label{Tab:variant}
		\begin{tabular}{l|cc}
			\hline
			Methods              & A$^2$LP              & A$^2$LP variant  \\
			\hline
			Acc. (\%)             & 79.3                 & 77.9          \\
			\hline
		\end{tabular}
	\end{minipage}
	\hspace{0.2cm}
	\begin{minipage}{0.48\linewidth}
		\centering
		\caption{Illustration of effects of the entropy-based instance weights (\ref{Equ:cate_centers}) in A$^2$LP based on a 50-layer ResNet.}
		\label{Tab:entropy_weight}
		\begin{tabular}{l|cc}
			\hline
			Methods                        & A$\to$W              & W$\to$A  \\
			\hline
			A$^2$LP                        &  87.7                 & 75.9          \\
			A$^2$LP ($w_i$=1, $\forall i$) &  87.4                 & 75.4                       \\
			\hline
		\end{tabular}
	\end{minipage}
\end{table}

\paragraph{\textbf{Effects of instance weighting in A$^2$LP}}
We investigate the effects of entropy-based instance weights in reliable virtual instances generation (\ref{Equ:cate_centers}) of A$^2$LP in this section. As illustrated in Table \ref{Tab:entropy_weight}, A$^2$LP improves over A$^2$LP ($w_i$=1, $\forall i$), where all unlabeled instances are weighted equally, supporting that instances of low entropy are more confident in terms of their predicted labels.

\begin{table}[h]
	\centering
	\caption{Results on the Office31 dataset \cite{saenko2010adapting} (ResNet-50).}
	\label{Tab:final_office31}
	\begin{tabular}{l|ccccccc}
		\hline
		%\hline
		Methods                          & A $\to$ W     & D $\to$ W     & W $\to$ D         & A $\to$ D     & D $\to$ A      & W $\to$ A     & Avg.  \\
		\hline
		Source Only                     & 68.4$\pm$0.2  & 96.7$\pm$0.1  & 99.3$\pm$0.1     & 68.9$\pm$0.2  & 62.5$\pm$0.3   & 60.7$\pm$0.3   & 76.1 \\
		DAN \cite{dan}                  & 80.5$\pm$0.4  & 97.1$\pm$0.2  & 99.6$\pm$0.1     & 78.6$\pm$0.2  & 63.6$\pm$0.3   & 62.8$\pm$0.2   & 80.4 \\
		%RTN \cite{rtn}                  & 84.5$\pm$0.2  & 96.8$\pm$0.1  & 99.4$\pm$0.1     & 77.5$\pm$0.3  & 66.2$\pm$0.2   & 64.8$\pm$0.3   & 81.6 \\
		DANN \cite{dann}     & 82.0$\pm$0.4  & 96.9$\pm$0.2  & 99.1$\pm$0.1     & 79.7$\pm$0.4  & 68.2$\pm$0.4   & 67.4$\pm$0.5   & 82.2 \\
		%ADDA \cite{adda}                & 86.2$\pm$0.5  & 96.2$\pm$0.3  & 98.4$\pm$0.3     & 77.8$\pm$0.3  & 69.5$\pm$0.4   & 68.9$\pm$0.5   & 82.9 \\
		%JAN-A\cite{jan}                  & 86.0$\pm$0.4  & 96.7$\pm$0.3  & 99.7$\pm$0.1     & 85.1$\pm$0.4  & 69.2$\pm$0.3   & 70.7$\pm$0.5   & 84.6 \\
		%MADA \cite{mada}                & 90.0$\pm$0.1  & 97.4$\pm$0.1  & 99.6$\pm$0.1     & 87.8$\pm$0.2  & 70.3$\pm$0.3   & 66.4$\pm$0.3   & 85.2 \\
		%SimNet \cite{SimNet}            & 88.6$\pm$0.5  & 98.2$\pm$0.2  & 99.7$\pm$0.2     & 85.3$\pm$0.3  & 73.4$\pm$0.8   & 71.8$\pm$0.6   & 86.2 \\
		% Kang \emph{et al.} \cite{attention_alignment}   & 86.8$\pm$0.2  & \textbf{99.3}$\pm$0.1  & \textbf{100.0}$\pm$.0     & 88.8$\pm$0.4  & 74.3$\pm$0.2   & 73.9$\pm$0.2   & 87.2 \\
		CDAN+E \cite{cada}               & 94.1$\pm$0.1  & 98.6$\pm$0.1  & \textbf{100.0}$\pm$.0     & 92.9$\pm$0.2  & 71.0$\pm$0.3   & 69.3$\pm$0.3   & 87.7 \\
		%TADA \cite{tada}                 & 94.3$\pm$0.3  & 98.7$\pm$0.1  & 99.8$\pm$0.2   & 91.6$\pm$0.3      & 72.9$\pm$0.2  & 73.0$\pm$0.3  & 88.4 \\
		SymNets \cite{symnets}           & 90.8$\pm$0.1  & 98.8$\pm$0.3           & \textbf{100.0}$\pm$.0 & 93.9$\pm$0.5 & 74.6$\pm$0.6                   & 72.5$\pm$0.5   & 88.4  \\
		DADA \cite{dada}                    & 92.3$\pm$0.1   & 99.2$\pm$0.1     & 100.0$\pm$0.0 & 93.9 $\pm$0.2 & 74.4$\pm$0.1 & 74.2$\pm$0.1 & 89.0 \\
		%TAT \cite{liu2019transferable}   & \textbf{92.5}$\pm$0.3  & \textbf{99.3}$\pm$0.1  & \textbf{100.0}$\pm$.0      & \textbf{93.2}$\pm$0.2  & 73.1$\pm$0.3   & 72.1$\pm$0.3  & 88.4 \\
		%MDD \cite{zhang2019bridging}     & 94.5$\pm$0.3  & 98.4$\pm$0.1  & \textbf{100.0}$\pm$.0   & 93.5$\pm$0.2     & 74.6$\pm$0.3  & 72.2$\pm$0.1   & 88.9 \\
		%CADA-P \cite{Kurmi_2019_CVPR}    & \textbf{97.0}$\pm$0.2  & \textbf{99.3}$\pm$0.1  & \textbf{100.0}$\pm$.0     & \textbf{95.6}$\pm$0.1  & 71.5$\pm$0.2            & 73.1$\pm$0.3            & 89.5 \\	
		CAN \cite{can}                   & \textbf{94.5}$\pm$0.3  & \textbf{99.1}$\pm$0.2  & 99.8$\pm$0.2     &95.0$\pm$0.3   &78.0$\pm$0.3    & 77.0$\pm$0.3   & 90.6 \\
		\hline
		LP                               & 81.1         & 96.8           & 99.0              & 82.3          & 71.6           & 73.1          & 84.0 \\
		A$^2$LP (ours)                   & 87.7         & 98.1           & 99.0              & 87.8          & 75.8           & 75.9          & 87.4 \\
		\hline
		MSTN (reproduced)                & 92.7$\pm$0.5 & 98.5$\pm$0.2   & 99.8$\pm$0.2      & 89.9$\pm$0.3  & 74.6$\pm$0.3   & 75.2$\pm$0.5  & 88.5 \\
		\quad empowered by A$^2$LP       & 93.1$\pm$0.2 & 98.5$\pm$0.1   & 99.8$\pm$0.2      & 94.0$\pm$0.2  & 76.5$\pm$0.3   & 76.7$\pm$0.3  & 89.8 \\
		\hline
		CAN (reproduced)                 & 94.0$\pm$0.5 & 98.5$\pm$0.1   & 99.7$\pm$0.1   & 94.8$\pm$0.4   & \textbf{78.1}$\pm$0.2 & 76.7$\pm$0.3 & 90.3   \\
		\quad empowered by A$^2$LP       & 93.4$\pm$0.3 & 98.8$\pm$0.1   & \textbf{100.0}$\pm$.0   & \textbf{96.1}$\pm$0.1   & \textbf{78.1}$\pm$0.1 & \textbf{77.6}$\pm$0.1 & \textbf{90.7}   \\
		%		\textbf{A$^2$LP + CE}            & 93.2$\pm$0.3  & \textbf{99.1}$\pm$0.1     & \textbf{100.0}$\pm$.0  & \textbf{95.4}$\pm$0.2  & 77.1$\pm$0.2 & \textbf{78.0}$\pm$0.1          & 90.5 \\
		\hline
	\end{tabular}
\end{table}

\begin{table}[h!]
	\centering
	\caption{Results on the ImageCLEF-DA dataset \cite{ImageCLEFDA} (ResNet-50). }
	\label{Tab:ImageCLEF}
	\begin{tabular}{l|ccccccc}
		\hline
		Methods                         & I $\to$ P     & P $\to$ I     & I $\to$ C         & C $\to$ I         & C $\to$ P      & P $\to$ C      & Avg.  \\
		\hline
		Source Only                     & 74.8$\pm$0.3  & 83.9$\pm$0.1  & 91.5$\pm$0.3     & 78.0$\pm$0.2  & 65.5$\pm$0.3   & 91.2$\pm$0.3   & 80.7 \\
		DAN \cite{dan}                  & 74.5$\pm$0.4  & 82.2$\pm$0.2  & 92.8$\pm$0.2     & 86.3$\pm$0.4  & 69.2$\pm$0.4   & 89.8$\pm$0.4   & 82.5 \\
		%RTN \cite{rtn}                  & 74.6$\pm$0.3  & 85.8$\pm$0.1  & 94.3$\pm$0.1     & 85.9$\pm$0.3  & 71.7$\pm$0.3   & 91.2$\pm$0.4   & 83.9 \\
		DANN \cite{dann}             & 75.0$\pm$0.6  & 86.0$\pm$0.3  & 96.2$\pm$0.4     & 87.0$\pm$0.5  & 74.3$\pm$0.5   & 91.5$\pm$0.6   & 85.0 \\
		%JAN \cite{jan}                  & 76.8$\pm$0.4  & 88.0$\pm$0.2  & 94.7$\pm$0.2     & 89.5$\pm$0.3  & 74.2$\pm$0.3   & 91.7$\pm$0.3   & 85.8 \\
		%MADA \cite{mada}                & 75.0$\pm$0.3  & 87.9$\pm$0.2  & 96.0$\pm$0.3     & 88.8$\pm$0.3  & 75.2$\pm$0.2   & 92.2$\pm$0.3   & 85.8 \\
		CDAN+E \cite{cada}              & 77.7$\pm$0.3  & 90.7$\pm$0.2  & \textbf{97.7}$\pm$0.3     & 91.3$\pm$0.3  & 74.2$\pm$0.2   & 94.3$\pm$0.3   & 87.7 \\
		%CADA-P \cite{Kurmi_2019_CVPR}   & 78.0          & 90.5          & 96.7             & 92.0          & 77.2           & 95.5           & 88.3  \\
		%TAT\cite{liu2019transferable}   & 78.8$\pm$0.2  & 92.0$\pm$0.2  & 97.5$\pm$0.3      & 92.0$\pm$0.3      & 78.2$\pm$0.4   & 94.7$\pm$0.4  & 88.9 \\
		SymNets \cite{symnets}          & \textbf{80.2}$\pm$0.3  & 93.6$\pm$0.2 & 97.0$\pm$0.3 & \textbf{93.4}$\pm$0.3 & 78.7$\pm$0.3  & 96.4$\pm$0.1  &89.9  \\
		\hline
		LP                              & 77.1         & 89.2           & 93.0             & 87.5               & 69.8           & 91.2          & 84.6 \\
		A$^2$LP (ours)                  & 79.3         & 91.8           & 96.3             & 91.7               & 78.1           & 96.0          & 88.9 \\
		\hline
		MSTN (reproduced)               & 78.3$\pm$0.2 &92.5$\pm$0.3    & 96.5$\pm$0.2     & 91.1$\pm$0.1 & 76.3$\pm$0.3&   94.6$\pm$0.4     & 88.2 \\
		\quad empowered by A$^2$LP      & 79.6$\pm$0.3 &92.7$\pm$0.3    & 96.7$\pm$0.1     & 92.5$\pm$0.2 & 78.9$\pm$0.2&   96.0$\pm$0.1     & 89.4 \\
		\hline
		CAN (reproduced)                 & 78.5$\pm$0.3           & 93.0$\pm$0.3 & 97.3$\pm$0.2 & 91.0$\pm$0.3          & 77.2$\pm$0.2      & \textbf{97.0}$\pm$0.2 & 89.0 \\
		\quad empowered by A$^2$LP       & 79.8$\pm$0.2           & \textbf{94.3}$\pm$0.3 & \textbf{97.7}$\pm$0.2 & 93.0$\pm$0.3          & \textbf{79.9}$\pm$0.1      & 96.9$\pm$0.2 & \textbf{90.3} \\
		%\textbf{A$^2$LP + CE}           & 79.0$\pm$0.2  & \textbf{94.0}$\pm$0.1  & 97.2$\pm$0.1  & \textbf{93.4}$\pm$0.3  & 78.3$\pm$0.1  & \textbf{97.2}$\pm$0.3          & \textbf{89.9} \\
		\hline
	\end{tabular}
\end{table}

\subsection{Results} \label{Sec:results}
We report the classification results on the Office-31 \cite{saenko2010adapting}, ImageCLEF-DA \cite{ImageCLEFDA}, and VisDA-2017 \cite{peng2017visda} datasets in Table \ref{Tab:final_office31}, Table \ref{Tab:ImageCLEF}, and Table \ref{Tab:Visda}, respectively. Results of other methods are either directly reported from their original papers if available or quoted from \cite{cada,lee2019drop}. Compared to classical methods \cite{dann,dan} aiming at domain-invariant feature learning, the vanilla LP generally achieves better results via the graph-based SSL principle, certifying the efficacy of the SSL principles in UDA tasks. Our A$^2$LP improves over the LP significantly on all three UDA benchmarks, justifying the efficacy of the introduction of virtual instances for UDA. Additionally, we reproduce the state-of-the-art UDA methods of Moving Semantic Transfer Network (MSTN) \cite{semantic_align} and Contrastive Adaptation Network (CAN) \cite{can} with the released codes \footnote{\url{https://github.com/Mid-Push/Moving-Semantic-Transfer-Network} \url{https://github.com/kgl-prml/Contrastive-Adaptation-Network-for-Unsupervised-Domain-Adaptation}}; by replacing the pseudo label generators of MSTN and CAN with our A$^2$LP, we improve their results noticeably and achieve the new state of the art, testifying the effectiveness of the combination of A$^2$LP and domain-invariant feature learning.

%To illustrate the generality of our A$^2$LP to various network structures, we also report in Table \ref{Tab:ImageCLEF_alexnet} the classification results on the ImageCLEF-DA \cite{ImageCLEFDA} dataset based on a AlexNet \cite{alexnet}. The state-of-the-art results certify the efficiency of our proposed A$^2$LP.

%\begin{table*}[!ht]
%	\centering
%	\caption{Comparison with state-of-the-art methods on the VisDA-2017 dataset \cite{peng2017visda} for UDA based on a 50-layer ResNet. $^*$SE report averaged predictions of $16$ augmentations of each image. The A$^2$LP reported is the degenerated variant detailed in Sec. \ref{Sec:analysis}. Full results are presented in the Appendix.}
%	\label{Tab:Visda}
%	\begin{tabular}{l|cccccc|cccc}
%		\hline
%		Methods & DAN \cite{dan} & RTN \cite{rtn} & DANN \cite{dann} & CDAN+E \cite{cada} &Lee  \cite{lee2019drop}  & SE$^{*}$ \cite{french2018selfensembling} & LP  &  \textbf{A$^2$LP} & \textbf{A$^2$LP + CE}\\
%		\hline
%		Acc. (\%)& 53.0          & 53.6           & 55.0             & 70.0               & 76.2  & 82.8    & 69.8                  &  78.7      & \textbf{85.3} \\
%		\hline
%	\end{tabular}
%\end{table*}

\begin{table}[h!]
	\centering
	\caption{Results on the VisDA-2017 dataset. The A$^2$LP reported is the degenerated variant detailed in Sec. \ref{Sec:analysis}. Full results are presented in the appendices. }
	\label{Tab:Visda}
	\begin{tabular}{l|c|c}
		\hline
		Methods                                             & Acc. Based on a ResNet50 & Acc. Based on a ResNet101  \\
		\hline
		Source Only                                         & 45.6                   & 50.8      \\
		DAN \cite{dan}                                      & 53.0                   & 61.1   \\
		%RTN \cite{rtn}                                      & 53.6                   & --     \\
		DANN \cite{dann}                                    & 55.0                   & 57.4  \\
		MCD \cite{mcd}                                      & --                     & 71.9   \\
		CDAN+E \cite{cada}                                  & 70.0                   & --   \\
		LPJT \cite{li2019locality}                                    & --                     & 74.0  \\
		DADA \cite{dada}                                               & --                     & 79.8 \\
		Lee \emph{et al.} \cite{lee2019drop}                & 76.2                   & 81.5  \\
		CAN \cite{can}                                                  & --                     & 87.2 \\
		\hline
		LP                                                  & 69.8                   & 73.9  \\
		A$^2$LP (ours)                                      & 78.7                   & 82.7 \\
		\hline
		MSTN (reproduced)                                   & 71.9                   & 75.2  \\
		\quad empowered by A$^2$LP                          & 81.5                   & 83.7      \\
		\hline
		CAN (reproduced)                                    & 85.6                   & 87.2 \\
		\quad empowered by A$^2$LP                          & \textbf{86.5}          & \textbf{87.6}     \\
		%		\textbf{A$^2$LP + CE}                               & \textbf{85.3}  \\
		\hline
	\end{tabular}
\end{table}

\section{Conclusion}
Motivated by the relatedness of problem definitions between UDA and SSL, we study the use of SSL principles in UDA, especially the graph-based LP algorithm. We analyze the conditions of affinity graph/matrix to achieve better propagation of true labels to unlabeled instances, and accordingly propose a new algorithm of A$^2$LP, which potentially improves LP via generation of unlabeled virtual instances. An empirical scheme of virtual instance generation is particularly proposed for UDA via a weighted combination of unlabeled target instances. By iteratively using A$^2$LP to get high-quality pseudo labels of target instances and learning domain-invariant features involving the obtained pseudo-labeled target instances, new state of the art is achieved on three datasets, confirming the value of further investigating SSL techniques for UDA problems.

\noindent \scriptsize \textbf{Acknowledgment.} This work is supported in part by the Guangdong R\&D key project of China (Grant No.: 2019B010155001), the National Natural Science Foundation of China (Grant No.: 61771201), and the Program for Guangdong Introducing Innovative and Enterpreneurial Teams (Grant No.: 2017ZT07X183). Correspondence to Kui Jia (emai: kuijia@scut.edu.cn)

\bibliographystyle{splncs04}
\bibliography{egbib}
\appendix

\section{Proof of Proposition 1}
\begin{proof}\label{PROOF-ACC}
	For any $\x_i, \x_j \in X$, we define the boolean value $P(\x_i,\x_j) := TRUE$ iff there exists a sequence of instances $(\x_i,\x_{t_1},\x_{t_2},...,\x_{t_m},\x_j)$ such that the product of their pair-wise similarities: $\a_{it_1}\cdot \a_{t_1t_2}\cdot...\cdot \a_{t_mj} \neq 0$. Then under the assumption made in Proposition \textcolor{red} {1}, minimizing Equation (\textcolor{red} {7}) results in $\hat{y_i} = \hat{y_j}$ for those with $P(\x_i,\x_j) = TRUE$, and $\hat{y_i} = y_i$ for all $\x_i\in X_L$, therefore we have
	\begin{equation} \tag{A.1}
	Acc = \frac{|\{\x_i\in X_U:\exists \x_j\in X_L, P(\x_i,\x_j)\}|}{u}.
	\end{equation}
	Obviously, enhancing the zero-valued similarity $\a_{mn}$ between a data instance $\x_m$ (labeled or unlabeled) and a labeled instance $\x_n \in X_L$, where $y_m = y_n$, to a positive number leads to non-decreasing value of $|\{\x_i\in X_U:\exists \x_j\in X_L, P(\x_i,\x_j)\}|$, and therefore non-decreasing value of $Acc$. In particular, if $\x_m \in X_U$ and originally $P(\x_m,\x_j) = FALSE, \forall \x_j \in X_L$, the prediction of $\x_m$ changes from original $\hat{y}_m(\neq y_m)$ to $y_n(=y_m)$ and thus the value of $Acc$ increases.
\end{proof}

\section{Analysis}

\paragraph{\textbf{Hyper-parameter $\alpha$ }} We investigate different values of $\alpha$ (of Equ. (\textcolor{red}{7})) in A$^2$LP. As illustrated in Table \ref{Tab:alpha}, the results are stable under a wide range of $\alpha$ (i.e., 0.1$\sim$0.75).
\begin{table}[h!]
	\centering
	\begin{tabular}{l|cccccccc}
		\hline
		Values of $\alpha$ & 0.1  & 0.25 & 0.4 & 0.5 & 0.6 & 0.75 & 0.9 & 2.0 \\
		\hline
		Acc. (\%) of A$^2$LP    & 95.7 & 96.2 & 96.2& 96.0& 96.0 & 95.8 & 94.3&16.8 \\
		\hline
	\end{tabular}
	\caption{Results of A$^2$LP with different values of $\alpha$ on the \textbf{P} $\to$ \textbf{C} task of the ImageCLEF-DA dataset.}
	\label{Tab:alpha}
\end{table}

\paragraph{\textbf{Practical Efficiency}}
To make the proposed methods applicable to datasets with large numbers of instances, we improve the dominating computations of our methods by adopting the NN-Descent \cite{dong2011efficient} to construct the $k$-nearest neighbor graph (\textcolor{red}{6}) and the conjugate gradient \cite{hestenes1952methods,zhu2005semi} to acquire the label predictions $\F^*$ (\textcolor{red}{11}). As illustrated in Table \ref{Tab:time-consuming}, the NN-Descent \cite{dong2011efficient} accelerates the brute-force implementation of affinity matrix at a factor of $20$, and the conjugate gradient-based solution (\textcolor{red}{11}) accelerates the closed-form solution (\textcolor{red}{7}) at a factor of $> 20$ on the VisDA-2017-Small task of $24K$ instances, while the classification results drop negligibly (in fact no drop at the precision level of $0.1\%$).

\begin{table}
	\caption{An illustration of the wall-clock time of the (a) graph construction (\textcolor{red}{6}) and (b) prediction solution (\textcolor{red}{7}) with different implementations on the VisDA-2017-Small task of $24K$ instances based on the Intel Xeon E5-2630 V4 CPU of $2.20$GHz.}
	\label{Tab:time-consuming}
	\begin{minipage}{0.48\linewidth}
		\begin{tabular}{l|cc}
			\hline
			Methods& Acc. (\%)& Time (s) \\
			\hline
			Brute-force impl. & 79.3 & 182 \\
			NN-Descent \cite{dong2011efficient}        & 79.3 & 9.0\\
			\hline
		\end{tabular}
		\subcaption{Graph construction (\textcolor{red}{6})}
	\end{minipage}
	\begin{minipage}{0.48\linewidth}
		\begin{tabular}{l|cc}
			\hline
			Methods &Acc. (\%) & Time (s) \\
			\hline
			Closed-form solution (\textcolor{red}{7}) & 79.3 &  51.2 \\
			CG \cite{hestenes1952methods,zhu2005semi} (\textcolor{red}{11})          & 79.3 &  2.4 \\
			\hline
		\end{tabular}
		\subcaption{Predictions solution (\textcolor{red}{7})}
	\end{minipage}
\end{table}

\section{Full Results of VisDA-2017}
The full classification results on the VisDA-2017 dataset are illustrated in Table \ref{Tab:visda-cate}.

\begin{table}[t]
	\centering
	\caption{Full classification results on the VisDA-2017 dataset for unsupervised domain adaptation (UDA).}
	\label{Tab:visda-cate}
	\begin{tabular}{l|cccccccccccc|c}
		\hline
		Methods                         & \rotatebox{90}{aero.}   & \rotatebox{90}{bike}   & \rotatebox{90}{bus}  & \rotatebox{90}{car}   & \rotatebox{90}{horse} & \rotatebox{90}{knife} & \rotatebox{90}{moto.}  & \rotatebox{90}{person} & \rotatebox{90}{plant} & \rotatebox{90}{sktb.} & \rotatebox{90}{train} & \rotatebox{90}{truck} & \rotatebox{90}{Avg.}  \\
		\hline
		\multicolumn{14}{l}{Results based on a 50-layer ResNet} \\
		\hline
		LP                                      & 91.4    & 81.4   & 73.3 & 71.8  & 94.7  & 60.8  & 87.4   & 62.2   & 87.8  & 19.1  & 86.2  & 20.9  & 69.8 \\
		A$^2$LP                                 & 95.5    & 82.8   & 77.9 & 70.0  & 95.2  & 95.9  & 86.6   & 65.3   & 87.4  & 42.8  & 86.4  & 53.1  & 78.7 \\
		\hline
		\tiny {MSTN (reproduced)}               & 86.9    & 73.2   & 76.8 & 67.2  & 80.7  & 78.8  & 71.9   & 65.1   & 74.8  & 76.2  & 85.6  & 25.6  & 71.9 \\
		\tiny {\quad empowered by A$^2$LP}      & 96.1    & 83.5   & 78.3 & 70.8  & 95.7  & 96.3  & 87.1   & 66.4   & 87.4  & 76.4  & 86.7  & 53.8  & 81.5 \\
		\hline
		\tiny {CAN (reproduced)}                & 94.5    & 85.4   & \textbf{81.9} & \textbf{72.3}  & 96.7  & 94.9  & 88.3   & 78.4   & \textbf{96.3}  & 94.7  & 86.2  & 57.3  & 85.6 \\
		\tiny {\quad empowered by A$^2$LP}      & \textbf{96.3}    & \textbf{86.2}   & 81.4 & 71.7  & \textbf{97.1}  & \textbf{96.8}  & \textbf{89.7}   & \textbf{79.1}   & 96.1  & \textbf{95.4}  & \textbf{88.6}  & \textbf{59.1}  & \textbf{86.5} \\
		
		\hline
		\hline
		\multicolumn{14}{l}{Results based on a 101-layer ResNet} \\
		\hline
		LP                                      & 89.6    & 80.6   & 65.4 & 72.9  & 92.7  & 74.0  & 84.2   & 72.8   & 87.9  & 48.4  & 84.6  & 33.0  & 73.9 \\
		A$^2$LP                                 & 96.0    & 82.9   & 82.2 & 68.9  & 95.8  & 96.0  & 87.8   & 66.5   & 89.6  & 85.2  & 88.4  & 53.2  & 82.7 \\
		\hline
		\tiny {MSTN (reproduced)}               & 90.5    & 73.0   & 70.2 & 58.9  & 84.9  & 77.0  & 84.5   & 79.3   & 89.6  & 69.6  & 89.4  & 36.0  & 75.2 \\
		\tiny {\quad empowered by A$^2$LP}      & 96.4    & 84.1   & 82.4 & 70.1  & 96.1  & 96.6  & 88.2   & 67.7   & 91.5  & 87.5  & \textbf{89.9}  & 54.0  & 83.7 \\
		\hline
		\tiny {CAN (reproduced)}                & 97.0    & \textbf{87.2}   & 82.5 & \textbf{74.3}  & 97.8  & 96.2  & \textbf{90.8}   & 80.7   & 96.6  & 96.3  & 87.5  & 59.9  & 87.2 \\
		\tiny {\quad empowered by A$^2$LP}      & \textbf{97.5}    & 86.9   & \textbf{83.1} & 74.2  & \textbf{98.0}  & \textbf{97.4}  & 90.5   & \textbf{80.9}   & \textbf{96.9}  & \textbf{96.5}  & 89.0  & \textbf{60.1}  & \textbf{87.6} \\
		\hline
	\end{tabular}
\end{table}

\end{document}